\documentclass{article}

\usepackage{nips_2017}
\usepackage{graphicx}

\usepackage[utf8]{inputenc} % allow utf-8 input
\usepackage[T1]{fontenc}    % use 8-bit T1 fonts
\usepackage{hyperref}       % hyperlinks
\usepackage{url}            % simple URL typesetting
\usepackage{booktabs}       % professional-quality tables
\usepackage{amsfonts}       % blackboard math symbols
\usepackage{nicefrac}       % compact symbols for 1/2, etc.
\usepackage{microtype}      % microtypography
\usepackage{amsmath, amssymb}
\usepackage{MarkMathCmds}
\usepackage{cancel}
\usepackage[utf8]{inputenc}
\usepackage{pgfplots}
\usepgfplotslibrary{groupplots}

\newcommand{\lb}{\mathcal{L}}

\title{Closed-form Inference and Prediction in \\ Gaussian Process State-Space Models}
\author{
  Alessandro Davide Ialongo \qquad Mark van der Wilk \qquad Carl Edward Rasmussen \\[5pt]
  Computational and Biological Learning Laboratory\\
  University of Cambridge\\
  Cambridge, UK\\
  \texttt{\{adi24, mv310, cer54\}@cam.ac.uk} \\
}

\begin{document}

\maketitle

\begin{abstract}
  We examine an analytic variational inference scheme for the Gaussian Process State Space Model (GPSSM) -- a probabilistic model for system identification and time-series modelling. Our approach performs variational inference over both the system states and the transition function. We exploit Markov structure in the true posterior, as well as an inducing point approximation to achieve linear time complexity in the length of the time series. Contrary to previous approaches, no Monte Carlo sampling is required: inference is cast as a deterministic optimisation problem.
  In a number of experiments, we demonstrate the ability to model non-linear dynamics in the presence of both process and observation noise as well as to impute missing information (e.g. velocities from raw positions through time), to de-noise, and to estimate the underlying dimensionality of the system. Finally, we also introduce a closed-form method for multi-step prediction, and a novel criterion for assessing the quality of our approximate posterior.
  
%  and uses an analytic lower bound on the model's marginal likelihood. By optimising this lower bound, we obtain an approximate 

%  We examine a fully variational method to fit Gaussian processes for system identification and time-series modelling. Inference in stochastic, non-linear dynamical systems is generally intractable, hence flexible approximations are required. Our approach performs variational inference over both the system states and the transition function, resulting in an analytic lower bound on the data’s marginal likelihood. We exploit the Markov structure of our posterior distribution as well as an inducing point approximation to achieve linear time complexity in the length of the time-series. Contrary to previous approaches, no Monte Carlo sampling is required: inference is cast as an optimisation problem.
% In a number of experiments, we demonstrate the ability to model arbitrary non-linear dynamics in the presence of both process and observation noise as well as to impute missing information (e.g. velocities from raw positions through time), to de-noise, and to estimate the underlying dimensionality of the system.

\end{abstract}

\section{Introduction}

The setting we consider is that of stochastic, non-linear dynamical systems under noisy measurements. Given discrete time data, recovering the system dynamics (i.e. a "transition function" from the current state to the next) can be likened to regression with noisy targets \emph{and} inputs. The state-space model formalism addresses this by distinguishing between observed variables and latent ones (the de-noised "states"). The system's evolution is assumed to be independent of the observations, given the states. Moreover, it is common to make a Markovian assumption and let the state at time \(t+1\) only depend on that at time \(t\) and not on any preceding ones: the system is fully specified by its current state.
% This is the well studied field of hidden Markov models \cite{baum1966statistical}.

In this work we focus on the Gaussian Process State Space Model (GPSSM), where the transition function is given a Gaussian process prior. We aim to perform approximate Bayesian inference both over the transition function and the unobserved states using a variational method. We follow on from work by \citet{frigola2014variational} and \citet{mchutchon2014nonlinear} and make three main contributions. Firstly we investigate the behaviour of a fully analytic variational inference scheme, and show how it performs on some initial tasks. Secondly, we introduce a new method for multi-step ahead prediction, based on augmenting the variational distribution. Finally, we also show how this method gives a way of assessing the quality of the approximate predictions and posterior at test-time, which was lacking in previous approaches.

%\section{Background}

%Contrary to the work of \cite{mattos2015recurrent}

\section{Doubly Variational Inference}
Here we briefly review a simplified derivation of variational inference in the GPSSM introduced by \citet{gpssmnote}. The main simplification over earlier presentations \citep{frigola2014variational,mchutchon2014nonlinear} is the use of a density over entire functions $q(f)$ (through a slight abuse of notation). While such a density does not strictly exist, we use it as an intermediate step before we integrate out all but a finite number of observations, at which point it becomes a familiar Gaussian distribution again.
% This greatly simplifies the derivation, as it removes the need to write out the distributions as incrementally larger conditionals.
We begin with the familiar form of the evidence lower bound (ELBO), with an approximate posterior over states $\vx$ and transition function $f$: $q(f, \vx)$. We constrain $q$ to be independent between $\vx$ and $f$, and we further use the sparse GP posterior introduced by \citet{titsias2009variational}.\footnote{See \citet{matthews2016sparse} for a more recent discussion.} We write $q(f)$ as the GP prior conditioned on a small set of \emph{inducing variables} $\vu$ together with a free density over $\vu$: $q(f|\vu)q(\vu)$. This gives a bound where the difficult GP conditional terms cancel out.
\begin{align}
\lb &= \Exp{q(f, \vx)}{\log\frac{p(\vx_0) \cancel{p(f|\vu)}p(\vu) \prod_{t=1}^T p(\vy_t|\vx_t) p(\vx_t|\vx_{t-1}, f)}{q(\vx)\cancel{p(f|\vu)}q(\vu)}} \\
&= \sum_{t=1}^T \Exp{q(\vx_t)}{\log p(\vy_t|\vx_t)} + \sum_{t=1}^T \Exp{q(x_{t-1:t})q(f)}{\log p(\vx_t|\vx_{t-1}, f)} + \Exp{q(\vx_0)}{\log p(\vx_0)} + \nonumber \\[5pt]
& \qquad \entropy{q(\vx)} - \KL{q(\vu)}{p(\vu)}
\end{align}
This general bound still allows a choice for the form of \(q(\vx)\) and \(q(\vu)\). \citet{frigola2014variational} samples from the intractable optimal variational distributions. We follow \citet{mchutchon2014nonlinear} by choosing a Gaussian \(q(\vx)\), which allows all expectations to be calculated in closed-form. Finally, the optimal \(q(\vu)\) is found by calculus of variations and is also a Gaussian with closed-form moments. We optimise the bound \(\lb\) with respect to: the GP hyper-parameters, the inducing inputs, the process noise standard deviations, the parameters of the linear-Gaussian emissions, and the moments of \(q(\vx)\).

\section{Predictions}
In addition to inference, making predictions in these models is itself challenging and also requires approximations or sampling. We will address the general case of multi-step ahead forecasting of a given sequence. The quality of existing methods for making approximate predictions can only be assessed using a held out test set. The variational framework, however, can be used to quantify the KL divergence between the exact predictive distribution, and the approximation. This provides a method to assess the reliability of the approximate predictive distribution, and the approximate posterior over states, without the need for a test set.

\paragraph{Exact predictions} After training on a sequence of data, we have a factorised approximate posterior for latent states $\vx$ and transition function $f$: $q(f)q(\vx)$. We can sample from the predictive distribution:
\begin{enumerate}
\item Sample from the final state's posterior $\vx_T \sim q(\vx_T)$ and the transition function's $f \sim q(f)$.\footnote{In practice, to sample $f$ we incrementally sample a function value, with the previous one serving as input.}
\item Incrementally sample future states $\vx_{T+p} \sim p(x_{T+p}|x_{T+p-1}, f)$ for $p \in \{1 \dots P\}$.
\item Sample predictive observations $\vy_t \sim p(\vy_t|\vx_t)$.
\end{enumerate}
This would be repeated in order to sample another trajectory. There are two main issues with the predictive distribution which make it difficult to deal with: a) there are correlations between $f$ and the states across all times, and b) the distributions become non-Gaussian after being passed through a GP several times. We will discuss two Gaussian approximate predictive distributions. 

\paragraph{Moment matching}
Given a distribution on an input of a GP $\vx \sim \NormDist{\mu, \Sigma}$, it is possible to analytically calculate the mean and variance of the GP output $f(\vx)$. This was used by \citet{girard2003multistep} to propose a moment matching approximation, where the moments of the next state are iteratively matched to those of the previous state passed through the GP. It was also used by \citet{deisenroth2011pilco} to make multi-step predictions for model based RL. The quality of this approximation degrades the further it has to predict, as any error introduced in the first moment matching step is passed on to the input of the next moment matching step.

\paragraph{Variational predictions}
Alternatively, we note that we can view the predictive distribution over states simply as a continuation of the posterior \(q(\vx)\) to states that do not have observations associated with them. Given this insight, we can augment the approximate posterior to include these extra states in the variational lower bound. This gives extra transition terms, without any terms from emission likelihoods. We can write the new augmented ELBO as
\begin{align}
\lb' = \lb + \sum_{p=T+1}^{T+P} \Exp{q(f, \vx)}{\log p\left(\vx_p | f, \vx_{p-1}\right) - \log q(\vx_p|\vx_{p-1})} \label{lb}
\end{align}
and maximise it w.r.t.~the added state distributions to make predictions, while keeping all the parameters obtained during training fixed. This gives an alternative Gaussian predictive distribution to moment matching.

\section{Assessing approximate predictions at test time}
We can expect both Gaussian approximations to do well in situations where the transitions are close to linear, or perhaps even when the state distribution remains unimodal. When state distributions become multimodal, which occurs when predicting near unstable equilibria in the dynamical system, we can not expect any Gaussian approximation to be close to the model's true predictive distribution. Here, we show that the augmented lower bound $\lb'$ can be used to monitor the accuracy of our approximate predictive distribution. We believe this to be important, since we can only expect to make good predictions if both our model is correct \emph{and} our predictions are consistent with our model.

To quantify the quality of our approximate predictive density, we will compute the KL divergence between it and the true predictive distribution. We start by considering the gap between the two bounds (where \(\vx^P\) are the predicted states):
\begin{align}
\lb - \lb' = \KL{q(\vx, f, \vx^P)}{p(\vx, f, \vx^P|\vy))} - \KL{q(\vx, f)}{p(\vx, f|\vy)}
\end{align}
We now note that the augmented KL divergence, is equal to the original one, plus terms depending on the predictive distributions, these terms (listed in (\ref{lb})) can also be written as:
\begin{align}
\Exp{q(f, \vx_T)}{\KL{q(\vx^P|\vx_T)}{p(\vx^P|f, \vx_T)}} = \lb - \lb'
\end{align}
This shows that the additional ELBO gap due to the additional states is equal to the expected KL divergence between the model's true predictive distribution over the states and the approximate one.

\section{Results and Conclusion}
\begin{figure}[h!]
	\centering
	\includegraphics[width=0.8\textwidth]{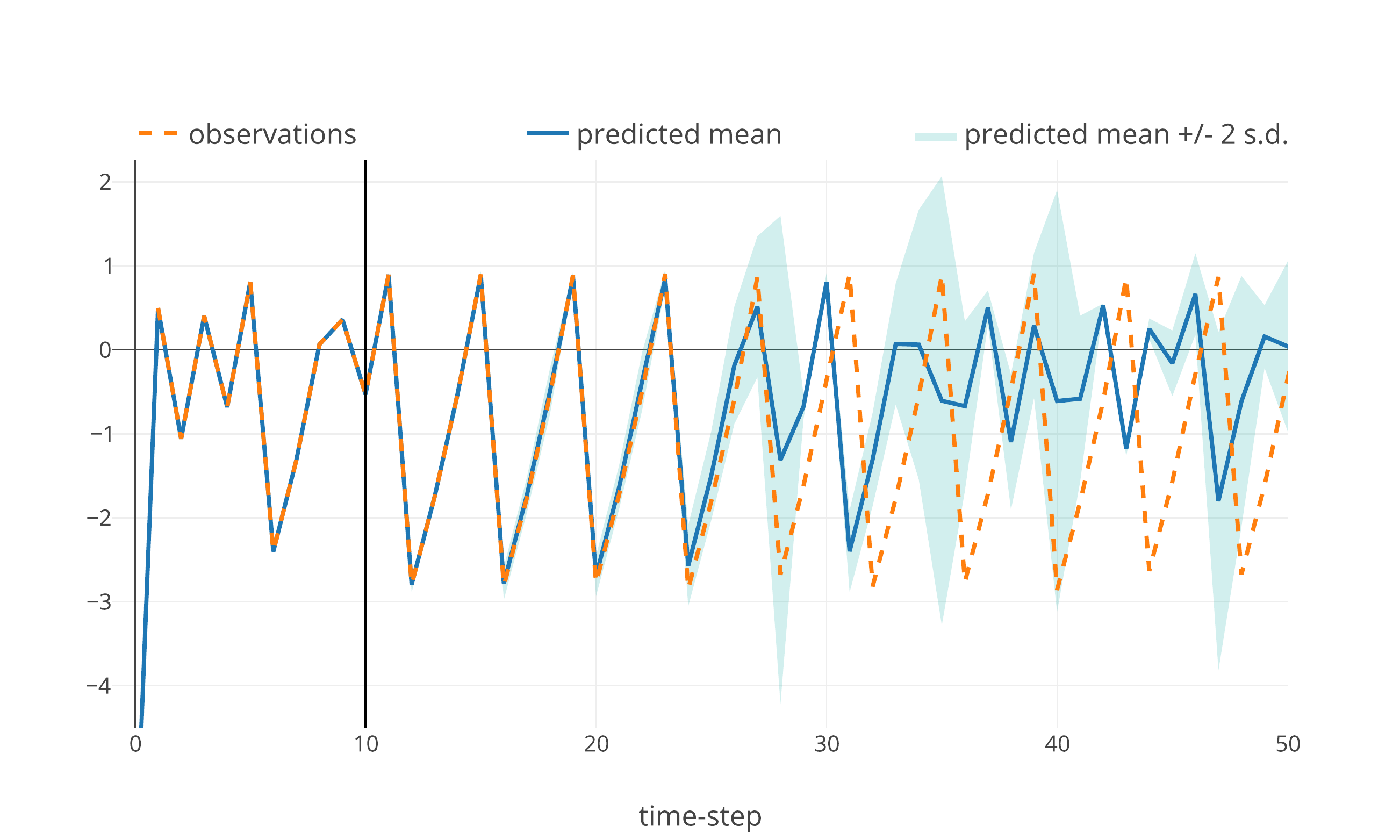}
    \caption{Kink dynamics: variational predictions. Model trained on near-noiseless data. Only data before the vertical black line is observed, in order to filter the latent states.}
    \label{fig:kink_var_preds}
\end{figure}

\begin{figure}[h!]
	\centering
	\includegraphics[width=\textwidth]{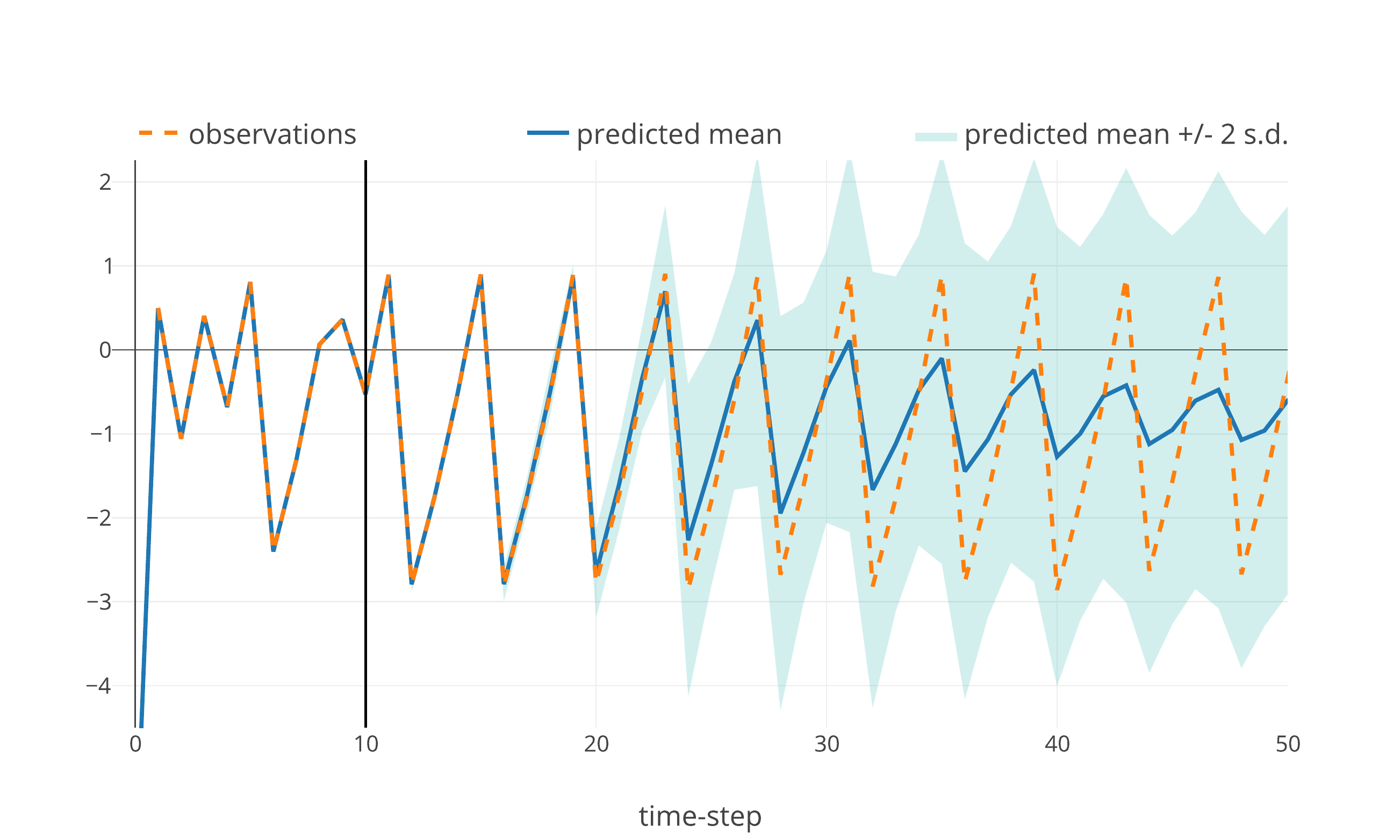}
    \caption{Kink dynamics: predictions obtained by sampling trajectories, using the same model as in Figure \ref{fig:kink_var_preds}.}
    \label{fig:kink_sampled_preds}
\end{figure}

\begin{figure}[h!]
  \centering
  \includegraphics[width=0.85\textwidth]{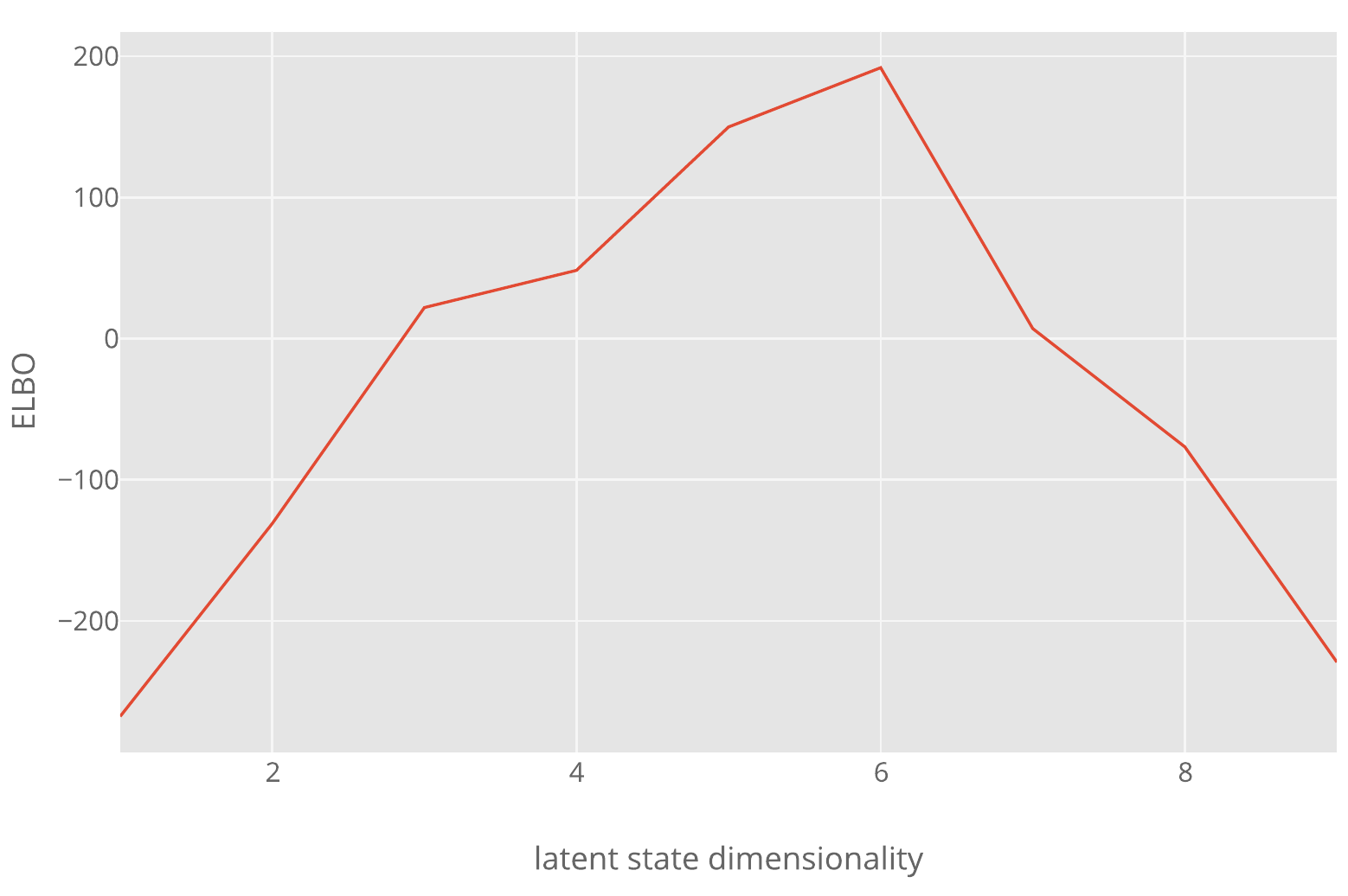}
  \caption{ELBO as a function of the latent state dimensionality. The model correctly identifies that a dimensionality of 5-6 is optimal (true dimensionality is 5) even though it only observes cart positions and the \(\sin\) and \(\cos\) of the pendulum angles. We initialise the surplus dimensions (if any) with the difference through time of positions and angles to help it model velocities (i.e. derivatives).}
  \label{fig:missing}
\end{figure}

We test out our model on two non-linear dynamical systems: the "kink" transition model (Figure \ref{fig:trans}) and the "Cart and Pole" (Figure \ref{fig:CP} in supplementary materials).
As we can see from Figures \ref{fig:trans}, \ref{fig:kink_var_preds} and \ref{fig:CP} the variational model can successfully learn the transition function from noisy data, filter on observed data (before the vertical black line), and predict multiple steps ahead. We see however that predictive performance degrades as we reach the non-linear region of the state-space where our Gaussian posterior is a rough approximation. This can be quantified by measuring the difference in bound degradation (Figure \ref{fig:degrad}) when predicting from different regions of the state-space. \\
Finally, in Figure \ref{fig:missing} we can see how the ELBO objective changes as a function of the model's latent dimensionality on the Cart and Pole dataset with omitted velocities. The ELBO objective has both a complexity penalty and a data-fit term which need to be traded off, leading it to be optimal close to the true dimensionality of the system (i.e. 5 latent dimensions), suggesting it is able to impute missing information. \\
Results under this variational inference scheme are promising, though they point out the fragility of a Gaussian \(q(\vx)\), especially when predicting variationally (compare with the more robust sampled predictions of Figures \ref{fig:CP} and \ref{fig:kink_sampled_preds}). Future work will explore the use of more flexible variational distributions.

% \begin{figure}[h!]
%   \centering
%   \begin{minipage}[b]{0.49\textwidth}
%   	\centering
%     \includegraphics[width=\textwidth]{gpssm_transition}
%     \caption{Learned transition function from noisy observations.}
%     \label{fig:trans}
%   \end{minipage}
%   \hfill
%   \begin{minipage}[b]{0.49\textwidth}
%   	\centering
%     \includegraphics[width=\textwidth]{kink_var_preds}
%     \caption{Variational predictions on near-noiseless data.}
%     \label{fig:kink_var_preds}
%   \end{minipage}
% \end{figure}

\begin{figure}[h!]
  	\centering
    \includegraphics[width=0.7\textwidth]{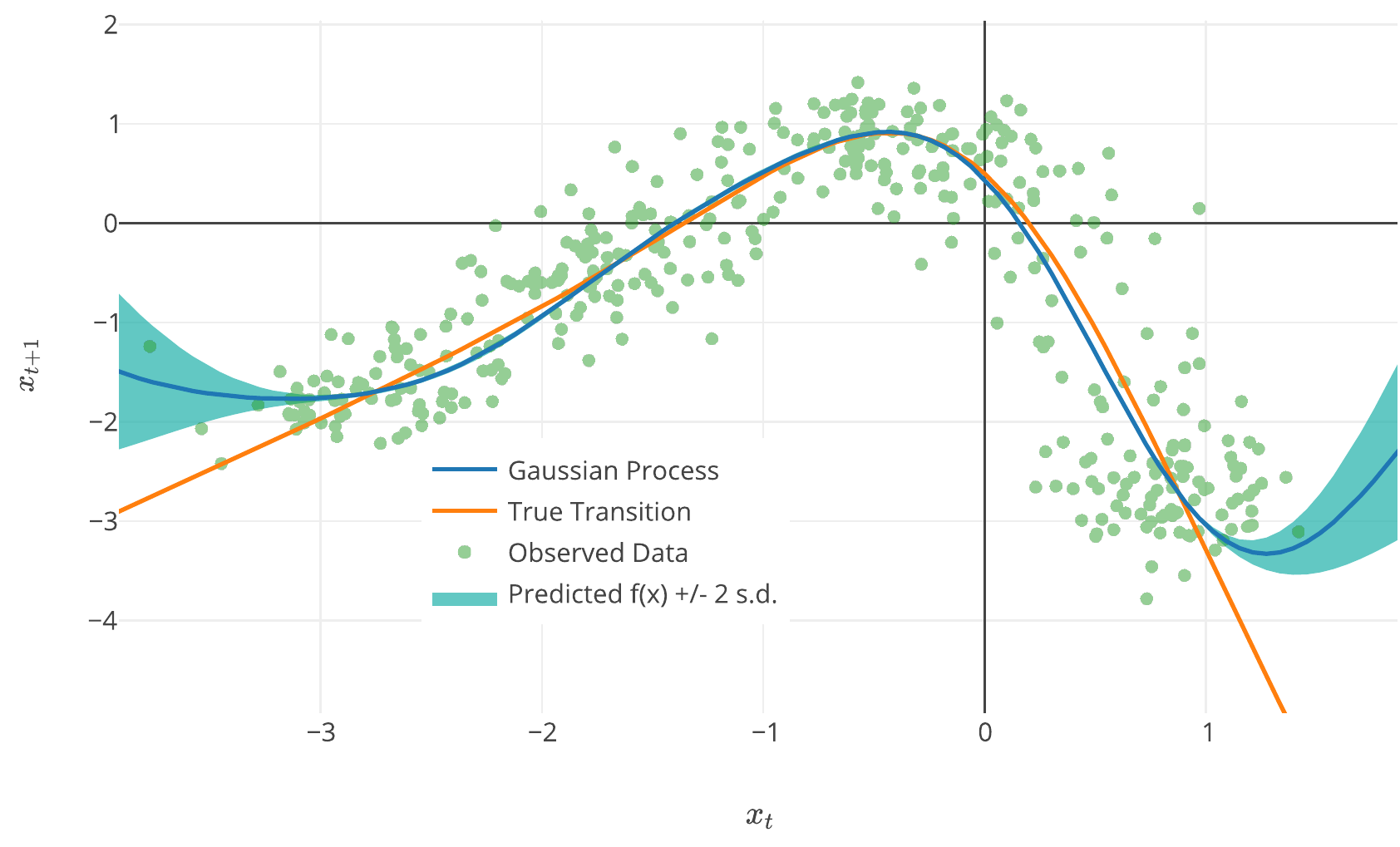}
    \caption{Kink dynamics: learned transition function from noisy observations.}
    \label{fig:trans}
\end{figure}

\begin{figure}[h!]
  \centering
%  \newlength\figureheight
%  \newlength\figurewidth
%  \setlength\figureheight{6cm}
%  \setlength\figurewidth{8cm}
  \scalebox{0.8}{\input{degrad.pgf}}
  \caption{Assessing the quality of the Gaussian predictive distribution in different regions of the state-space.
  \newline \textbf{Left:} linear regime, initial state: -2.5, degradation in ELBO due to prediction: -0.220 \newline \textbf{Right:} non-linear regime, initial state: -0.5, degradation in ELBO due to prediction: -46.9}
  \label{fig:degrad}
\end{figure}

\subsubsection*{Acknowledgments}
We would like to acknowledge Richard E. Turner, James Hensman and Hong Ge for helpful discussions. ADI and MvdW are generously supported by Qualcomm Innovation Fellowships.

\clearpage
\bibliography{references}

\newpage
\section{Supplementary Material}

\begin{figure}[h!]
  \centering
  \begin{minipage}[b]{\textwidth}
  	\centering
    \includegraphics[width=0.72\textwidth]{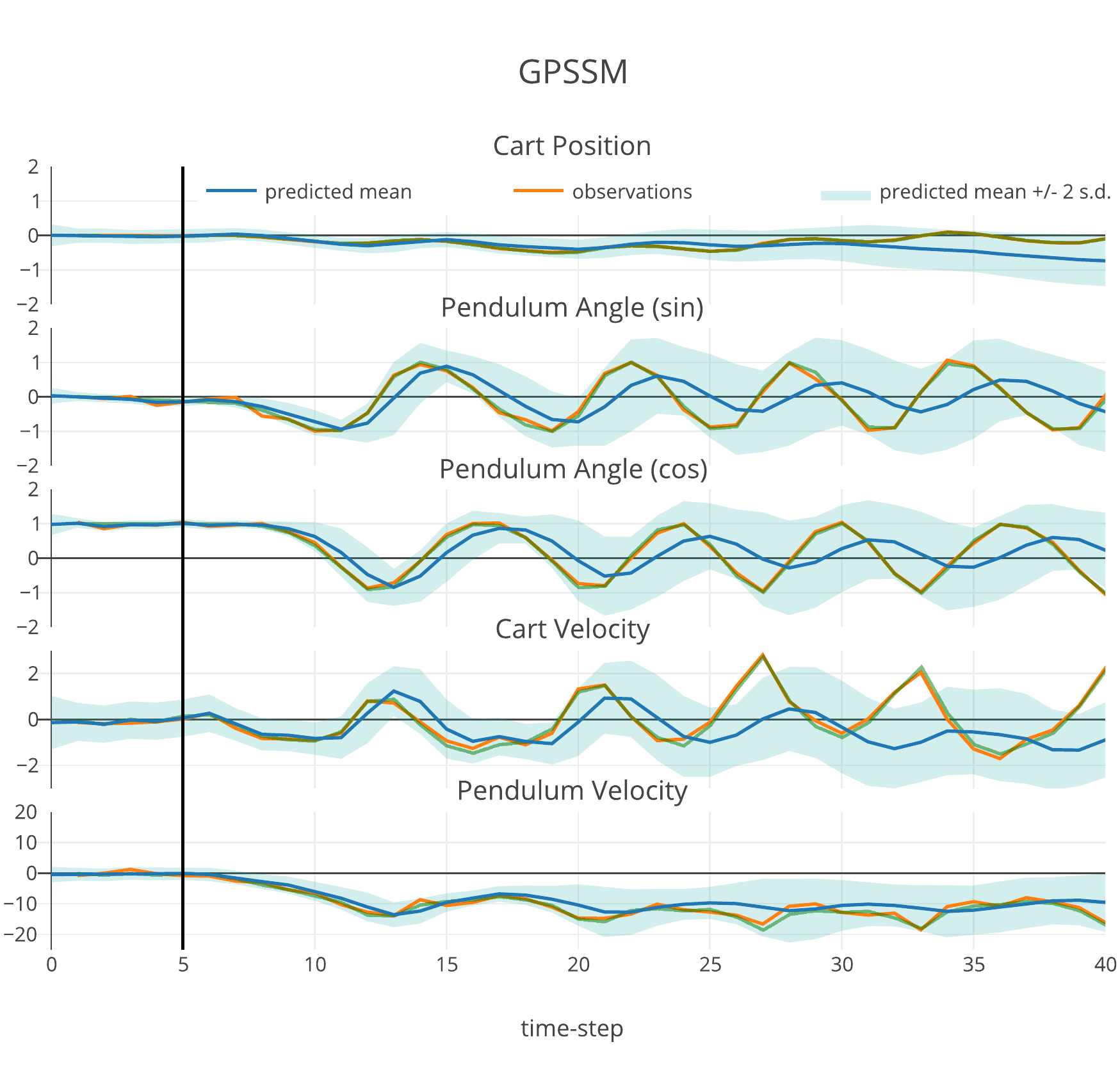}
  \end{minipage}
  \hfill
  \begin{minipage}[b]{\textwidth}
  	\centering
    \includegraphics[width=0.72\textwidth]{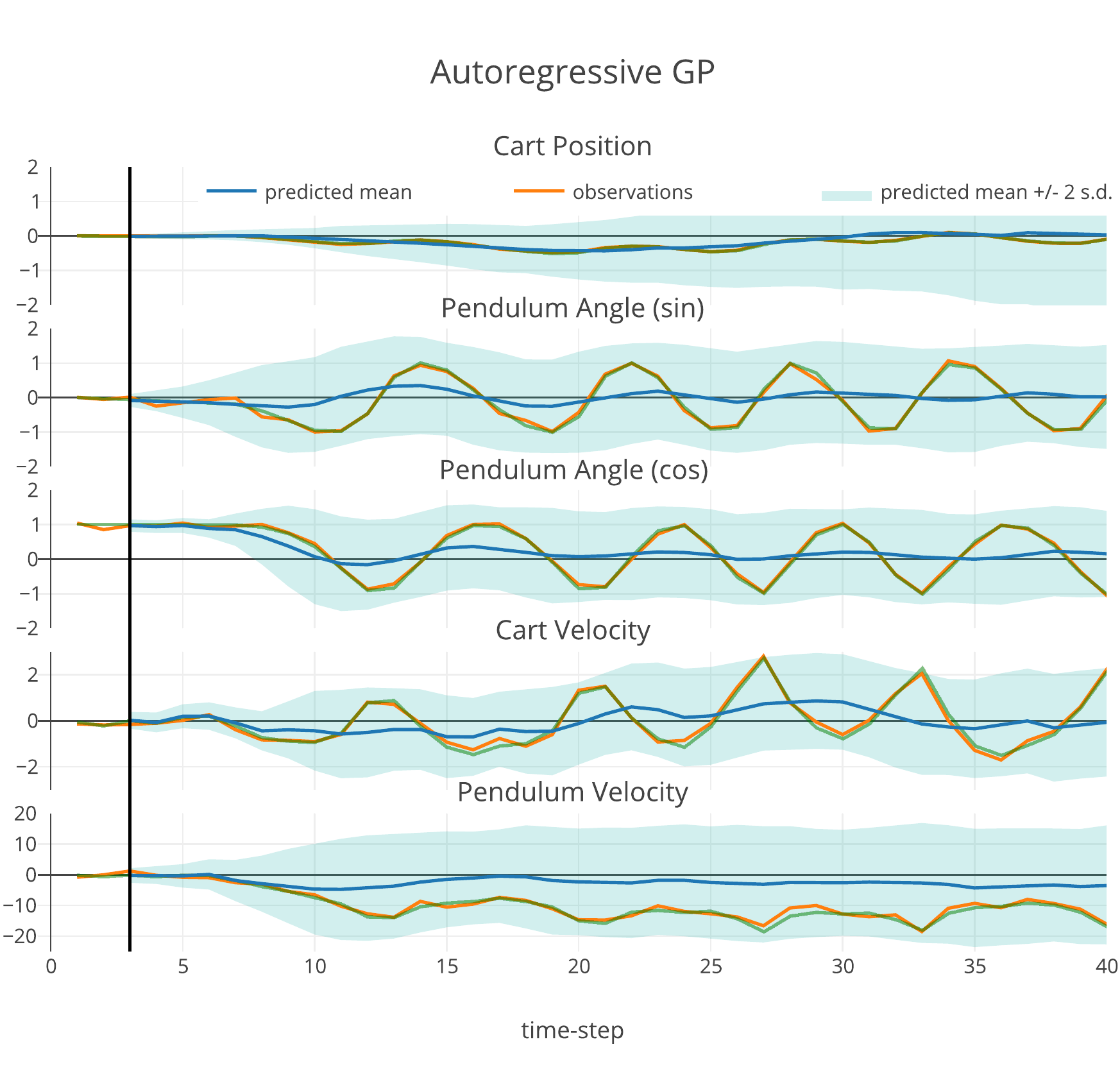}
  \end{minipage}
  \caption{Open-loop predictions for the cart and pole dynamical system. Control inputs are given and predictions are evaluated by sampling trajectories. Predictive performance is significantly improved by de-noising in the GPSSM. The true noiseless states (green line) are given for reference. Prediction begins after the black vertical line, the initial sequence is given to filter the state.}
  \label{fig:CP}
\end{figure}

\end{document}